\pdfoutput=1

\documentclass[11pt]{article}

\usepackage[]{EACL2023}

\usepackage{times}
\usepackage{latexsym}
\usepackage{multirow}
\usepackage{amsmath}
\usepackage{graphics}
\usepackage{graphicx}
\usepackage[T1]{fontenc}

\usepackage[utf8]{inputenc}

\usepackage{microtype}
\usepackage{amsmath}
\DeclareMathOperator{\GRL}{GRL}

\newcommand\scalemath[2]{\scalebox{#1}{\mbox{\ensuremath{\displaystyle #2}}}}

\usepackage{inconsolata}

\title{Zero-shot Transfer of Article-aware Legal Outcome Classification for European Court of Human Rights Cases}

\author{ Santosh T.Y.S.S$^{1}$, Oana Ichim$^2$, Matthias Grabmair$^1$
\\ $^1$School of Computation, Information, and Technology; \\ Technical University of Munich, Germany \\
$^2$Graduate Institute of International and Development Studies, Geneva, Switzerland\\ 
 \texttt{\{santosh.tokala, matthias.grabmair\}@tum.de} \\
  \texttt{oana.ichim@graduateinstitute.ch}
  }

\begin{document}

\maketitle

\begin{abstract}
In this paper, we cast Legal Judgment Prediction on European Court of Human Rights cases into an article-aware classification task, where the case outcome is classified from a combined input of case facts and convention articles. This configuration facilitates the model learning some legal reasoning ability in mapping article text to specific case fact text. It also provides an opportunity to evaluate the model's ability to generalize to zero-shot settings when asked to classify the case outcome with respect to articles not seen during training. We devise zero-shot experiments and apply domain adaptation methods based on domain discrimination and Wasserstein distance. Our results demonstrate that the article-aware architecture outperforms straightforward fact classification. We also find that domain adaptation methods improve zero-shot transfer performance, with article relatedness and encoder pre-training influencing the effect.

\end{abstract}

\section{Introduction}
\label{sec_intro}
Legal Judgment Prediction (LJP) has recently gained considerable attention in the mainstream NLP community (e.g., \citealt{aletras2016predicting, chalkidis2019neural,chalkidis2021paragraph, chalkidis2022lexglue,santosh2022deconfounding,santosh2023leveraging}). In LJP, the outcome of a case should be classified/predicted based on a textual description of case facts. In actual legal reasoning, legal practitioners (e.g., advocates, judges) determine relevant rules from the sources of law (e.g., statutes, regulations, precedent) that are relevant to the case at hand. They then carry out an analysis to determine which rules apply to the case at hand, and deduce the outcome of the case by applying them. Subsuming case facts under elements of rules given in legal sources plays a critical role in this process. Many current LJP approaches (e.g., \citealt{aletras2016predicting, chalkidis2019neural, chalkidis2022lexglue, santosh2023leveraging}) tackle this as a straightforward classification problem with the textual descriptions of case fact as the sole input. This reliance on the model learning statistical correspondences from case fact descriptions directly to outcomes neglects the role of legal sources in this relationship. As a consequence, the model may learn sub-optimal fact-outcome patterns that are informed by the case distribution in the data rather than learning to align facts with the legal source text containing applicable rules. The models may also attend to outcome-correlating distractors present in the dataset rather than engage in the legal fact-vs-law reasoning that is required of legal practitioners for a proper justification of the outcome \cite{santosh2022deconfounding}.

This work seeks to remedy this incomplete inference and enable the model to learn more authentic reasoning between rules and case facts by casting LJP into an article-aware classification setting and subjecting it to a zero-shot transfer challenge. Article-aware classification has been explored on Chinese criminal case corpora \cite{wang2018modeling, wang2019hierarchical, yue2021neurjudge, chen2022mulan}. Similarly,  \citealt{holzenberger2020dataset} has modeled statutory reasoning by classifying US tax law provisions concatenated with textual case descriptions. We build on this prior work in two ways. First, we develop and evaluate our model on a public dataset \cite{chalkidis2022lexglue} of cases by the European Court of Human Rights (ECtHR), which hears complaints by individuals about possible infringements of their rights enshrined in the European Convention on Human Rights (ECHR) by states. To the best of our knowledge, this is the first work applying  article-aware case outcome prediction setting to human rights adjudication. Our approach pairs case fact descriptions with candidate ECHR articles and assigns a binary target label depending on whether the article has been alleged/deemed to have been violated, or not. 
Our results show that the article-aware classification model outperforms the traditional classification setup by a small but consistent margin.

Second, we subject the model to a zero-shot transfer task. Models trained on case facts alone cannot produce inferences about convention articles they did not observe during training. By contrast, human judges can conduct outcome analysis with new/amended legal provisions because they are trained to understand the rules they contain and apply them to case facts in an expertise-informed way, even in the absence of secondary sources (e.g., commentaries to the rule, etc.). Article-aware classification allows an emulation of this process by means of a zero-shot benchmarking task on articles unseen at training time. We compare two conditions where (1) the model either has no access to the target articles, or (2) it is allowed to `read' the target articles but is not given any prediction outcome labels for case-target article pairs.

We experiment with domain adaptation by means of a domain discriminator \cite{ganin2016domain} and Wasserstein distance \cite{shen2018wasserstein}. Our results show that this improves performance on unseen articles compared to a vanilla model. We study the impact of law-specific pre-trained encoders on this zero-shot transferability compared to the standard language pre-trained one. Intuitively, we observe that our models perform better in zero-shot transfer if the target/unseen articles are semantically related to articles seen at training time.

It should be noted that, despite these tasks being typically referred to as instances of `legal judgment prediction', ECtHR fact statements are typically not finalized until the decision outcome is known, making the task effectively one of retrospective classification rather than prediction \cite{medvedeva2021automatic}. While this does lead to distracting and confounding phenomena (see our prior work in \citealt{santosh2022deconfounding}), the dataset remains a useful resource for the development of NLP models that analyze these fact statements for text patterns that correspond to specific convention articles as drafted by the court. Consequently, in this paper we hence speak of our models as engaging in \textit{case outcome classification (COC)}.

Our main contributions in this paper are\footnote{Our code is available at https://github.com/TUMLegalTech/zeroshotLJP}:
\begin{itemize}
    \item We cast LJP/COC on ECtHR cases as an article-aware classification task by pairing case fact descriptions with candidate articles. Assuming a frozen pre-trained encoder network, our article-aware prediction model outperforms straightforward fact classification.
    \item We conduct zero-shot transfer benchmarking of article-aware COC models. We find this to be a difficult testing task for the generalization of COC models. We show that domain adaptation using a domain discriminator and a Wasserstein distance method improves generalization.
    \item We conduct auxiliary experiments validating that article relatedness positively affects transfer performance and show an interaction between domain adaptation and domain specific encoder pre-training.

\end{itemize}

\section{Related Work}
\textbf{Legal Judgement Prediction:} LJP/COC as an NLP task has been studied using corpora from different jurisdictions, such as the ECtHR \cite{chalkidis2019neural, chalkidis2021paragraph, chalkidis2022lexglue, aletras2016predicting,liu2017predictive,  medvedeva2020using, says2020prediction, medvedeva2021automatic,santosh2023leveraging}
Chinese Criminal Courts \cite{luo2017learning,zhong2018legal, yang2019legal, yue2021neurjudge, zhong2020iteratively}, US Supreme Court \cite{katz2017general,kaufman2019improving}, Indian Supreme Court \cite{malik2021ildc,shaikh2020predicting}
the French court of Cassation \cite{csulea2017predicting, csulea2017exploring}, Brazilian courts \cite{lage2022predicting}, 
the Federal Supreme Court of Switzerland \cite{niklaus2021swiss}, 
UK courts \cite{strickson2020legal} and German courts \cite{waltl2017predicting}

Early works \cite{aletras2016predicting, csulea2017exploring,  csulea2017predicting, virtucio2018predicting, shaikh2020predicting, medvedeva2020using} used bag-of-words features. More recent approaches use deep learning \cite{zhong2018legal, zhong2020iteratively, yang2019legal}. Large pre-trained transformer models have since become the dominant model family in COC/LJP \cite{chalkidis2019neural,niklaus2021swiss}, including legal-domain specific pre-trained variants \cite{chalkidis2020legal,zheng2021does} that have been employed for the benchmark ECtHR corpus we use in this paper (\citealt{chalkidis2021paragraph, chalkidis2022lexglue}).

Prior work on Chinese criminal case corpora case extends fact-based classification by providing the text of legal source articles as additional input. \citealt{luo2017learning} used an attention-based neural network which jointly models  charge prediction and relevant article extraction in a unified framework whose input includes the text of legal articles. Similarly, \citealt{wang2018modeling,wang2019hierarchical,chen2022mulan,yue2021neurjudge} employ matching mechanism between case facts and article texts. To the best of our knowledge, ours is the first work to adapt article-aware prediction to the ECtHR corpus, which is situated in the in human rights litigation domain. Going beyond previous works, we further benchmark the zero-shot transfer performance of such models, providing a test bed to evaluate their capability to process  article texts they have not seen during training time and applying them to case facts towards classifying allegations/outcomes.


\noindent \textbf{Domain Adaptation (DA):} 
In transfer learning, the field of domain adaptation (DA) addresses the covariate shift between source and target data distributions \cite{ruder2019neural}. It is tackled under three different settings: (1) Semi-supervised DA \cite{bollegala2011relation, daume2006domain} where labels for the source and a small set of labels for the target domain are available, (2) unsupervised DA \cite{ganin2016domain, blitzer2006domain} where only labels for the source domain and unlabelled target data are given, and (3) Any Domain Adaptation / Out of Distribution generalization \cite{ben2021pada, volk2022example} where only labeled source data is given. In this work, we distill the existing public LexGLUE ECtHR dataset into a new benchmark on more challenging unsupervised and any domain adaptation settings for COC to emulate legal reasoning involving previously unseen convention articles. 

DA variants have been benchmarked for various NLP tasks, such as Question answering \cite{yu2018modelling}, duplicate question detection \cite{shah2018adversarial}, sentiment analysis \cite{li2017end, ganin2016domain}, dependency parsing \cite{sato2017adversarial}, relation extraction \cite{wu2017adversarial}, POS tagging \cite{yasunaga2018robust}, named entity recognition \cite{jia2019cross}, event trigger identification \cite{naik2020towards}, machine reading comprehension \cite{wang2019adversarial}, and machine translation \cite{yang2018unsupervised}. To the best of our knowledge, this work is the first to benchmark domain adaptation in COC/LJP. While previous works typically involve short text, COC on ECtHR data involves case facts and articles, both of which typically are long documents.

Methods proposed for domain adaptation can be categorized into four types: (a) Instance-based data selection methods \cite{jiang2007instance, remus2012domain} which employ similarity metrics to sample source data points to match the distribution of the target domain and train models based on obtained subsamples from the source domain, (b) Pseudo-labeling approaches \cite{ruder2018strong, rotman2019deep} which train a classifier based on source data initially and use it to predict labels on unlabeled target data towards further adapting the model, (c) Pivot-based methods \cite{blitzer2006domain,ziser2017neural} which aim to map different domains to a common latent space (where the feature distributions are close) by employing auto encoders and structural correspondence learning, and (d) Loss-based methods \cite{ganin2015unsupervised, shen2018wasserstein} which employ domain adversaries aiming to minimize the discrepancies between source and target data distributions. 
In this woork, we employ loss-based approaches using a domain discriminator \cite{ganin2016domain} and Wasserstein distance \cite{shen2018wasserstein} to enable domain adaptation for our COC models.

\section{Dataset, Tasks \&  Settings}
\label{sec_settings}
We use the LexGLUE ECtHR dataset provided by \cite{chalkidis2022lexglue}, which consists of 11k case fact descriptions along with target label information about which convention articles have been alleged to be violated (task B), and which the court has eventually found to have been violated (task A). The dataset is chronologically split into training (2001–2016), validation (2016–2017), and test set (2017-2019) with 9k, 1k, and 1k cases, respectively. The label set includes 10 prominent ECHR articles, which forms a subset of all the rights contained in the convention and its protocols. In both the ECtHR A and B benchmarks, it is assumed that the model classified the target from the fact description alone, which we refer to as the \textbf{fact classification variant}. 

For our article-aware classification settings, we augment the dataset with the texts of the 10 articles copied from the publicly available ECHR convention document\footnote{\url{https://www.echr.coe.int/documents/convention\_eng.pdf}}. We formulate the \textbf{article-aware prediction variant} for both tasks: Given both the case fact statements and a particular article information, the model should classify the binary outcome of whether an article has been alleged to be violated by the claimant (task B) or found to have been violated by the court (task A).

Our zero-shot transfer task then involves determining violation/allegation from case facts with respect to articles which are not seen during training time. We consider a `domain' to be a particular convention article (i.e., 10 convention articles form 10 domains). The objective is to train a model on a source domain (seen articles) with the goal of performing well at test-time on a target domain (unseen articles). Following \cite{yin2019benchmarking, ramponi2020neural}, we propose two settings under zero-shot COC:

\noindent \textbf{Zero-Shot Restrictive / Unsupervised Domain Adaptation (UDA):} In this setting, the model is given a pair of case facts and the text of training set articles (i.e., the source domain) along with their corresponding violation/allegation outcome label. In the target domain, it is provided with case facts and article text pairs as well, but the outcome label is withheld. 
The goal of UDA is to learn an outcome classifier from the outcome labelled source domain which should generalize well on the target domain by leveraging outcome-unlabeled target data. This setting is legally realistic, as the text of new or modified written legal sources is typically known for a given task and available for domain adaptation (e.g., a public administration decision support tool receives an update after relevant legislation has changed). 

\noindent \textbf{Zero-shot Wild / Any Domain Adaptation (ADA) / Out of Distribution Generalization:} In this setting, the model  never sees any article data from the target domain during training, yet should be able to generalize to it. In the legal setting, this corresponds to a model which is required to work with texts of sources only available at query time (e.g., complex retrieval settings where multiple legal sources potentially apply).

We reorganize the dataset to evaluate our zero-shot transfer/adaptation models by splitting the 10 ECHR articles into two non-overlapping groups, such that both contain articles of various frequencies (common, moderate, rare). 
\begin{itemize}
    \item \textbf{$split\_0$:} 6, 8, P1-1, 2, 9 
    \item \textbf{$split\_1$:} 3, 5, 10, 14, 11
\end{itemize}

\noindent We evaluate UDA and ADA on $split\_0$ as source and $split\_1$ as target, and vice-versa. 

\section{Method}
We employ a hierarchical neural model which takes the case fact description $x$ along with the article $a$ as input and outputs a binary outcome (allegation in Task B and violation in Task A) for case $x$ with respect to article $a$. Our architecture is a modified version of the Enhanced Sequential Inference Model (ESIM) \cite{chen2017enhanced} incorporating conditional encoding \cite{augenstein2016stance,rocktaschel2015reasoning} that has been adapted to deal with long input sequences following hierarchical attention networks \cite{yang2016hierarchical}. We experiment with two domain adaptation components based on adversarial training: (1) a classification-based domain discriminator and (2) a Wasserstein-distance based method which aims to reduce the difference between the source and the target domain distributions.

\subsection{Article-aware prediction Model}
\begin{figure}[]
\centering
 \scalebox{.5}{
\includegraphics[width =\textwidth,angle =90]{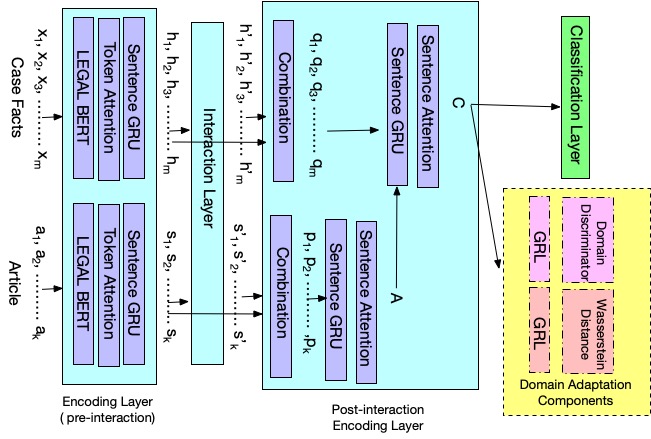}}
\caption{ Our article-aware prediction model  architecture}
\label{arch}
\end{figure}

Given the facts of the case $x = \{x_1, x_2, \ldots, x_m\}$ where $x_i = \{x_{i1}, x_{i2},\ldots,x_{in}\}$  and the article $a = \{a_1, a_2, \ldots, a_k\}$ where $a_j = \{a_{j1}, a_{j2},\ldots,a_{jl}\}$, the model outputs a binary label. $x_i$ / $a_i$ and $x_{jp}$ / $a_{jp}$ denote the $i^\text{th}$ sentence and $p^\text{th}$ token of the $j^\text{th}$ sentence of the case facts / article, respectively. $m / k$ and $n / l$ denote the number of sentences and tokens in the $i^\text{th}$ sentence of case facts / article, respectively. Our model contains an encoding layer, followed by an interaction layer, a post-interaction encoding layer, and a classification header. See Fig. \ref{arch} for an overview of our architecture. 

\subsubsection{Pre-interaction Encoding Layer} Our model encodes the facts of the case $x$ sentence-wise with LegalBERT \cite{chalkidis2020legal} to obtain token level representations $\{z_{i1}, z_{i2},\ldots, z_{in}\}$. These are aggregated into sentence level representations using token attention: 
\begin{equation}
\scalemath{0.8}{
    u_{it} = \tanh(W_w z_{it} + b_w ) } 
\end{equation}
\begin{equation}
\scalemath{0.8}{
    \alpha_{it} = \frac{\exp(u_{it}u_w)}{\sum_t \exp(u_{it}u_w)}  ~~\& ~~ 
    f_i = \sum_{t=1}^n \alpha_{it}z_{it}}
\label{att}
\end{equation}
where $W_w$,$b_w$ and $u_w$ are trainable parameters. The sentence level representations $\{f_1, \ldots, f_n\}$ are passed through a GRU encoder to obtain context-aware sentence representations of the facts $h = \{h_1, h_2, \ldots, h_m\}$. The analogous article encoder takes $a$ as input and outputs $s = \{s_1, s_2, \ldots, s_k\}$.

\subsubsection{Interaction Layer} Interaction between the sentences of the case facts and articles is done via dot product attention between the two sequences of sentences as follows: 

\begin{equation}
\scalemath{0.8}{
    e_{ij} = h_i^Ts_j ~~\&~~  h_i^\prime = \sum_{j=1}^{k}\frac{\exp(e_{ij})}{\sum_{l=1}^{k}\exp(e_{il})} s_j
}
\end{equation}
\begin{equation}
\scalemath{0.8}{
s_j^\prime = \sum_{i=1}^{m}\frac{\exp(e_{ij})}{\sum_{l=1}^{m}\exp(e_{lj})} h_i}
\end{equation}
where $e_{ij}$ represents the dot product interaction score between the context-aware representations of the $i^\text{th}$ sentence of the case facts and the $j^\text{th}$ sentence of the article. $h_i^\prime$ and $s_j^\prime$ represent article-aware representations corresponding to the $i^\text{th}$ sentence of the case facts and the fact-aware representation corresponding to the $j^\text{th}$ sentence of the article, respectively. Finally, we obtain  interaction-aware sentence representations of the facts $h^\prime = \{h^\prime_1, h^\prime_2, \ldots, h^\prime_m\}$. Similarly for the article, we obtain $s^\prime = \{s^\prime_1, s^\prime_2, \ldots, s^\prime_k\}$

\subsubsection{Post-Interaction Encoding Layer}
The article-dependent final representation of 
 the case facts is obtained in two steps: (i)  we compute the final representation of the article text  and (ii) use it as a conditional encoding \cite{augenstein2016stance,rocktaschel2015reasoning} to obtain the final article-dependent fact representation.

\noindent \textbf{Final representation of article}: We first combine the pre-interaction sentence encodings and fact-aware sentence representations of the article:
\begin{equation}
\scalemath{0.8}{
    p_i = [s_i, s_i^\prime, s_i - s_i^\prime, s_i \odot s_i^\prime]
    }
\label{eq_merge}
\end{equation}
where $\odot$ denotes element-wise product. This  representation aims to capture high-order interaction between the pre- and post- interaction elements \cite{chen2017enhanced}. The sentence representations $p_i$ are passed over a non-linear projection and  a GRU (as in the pre-interaction encoder) to perform context-level modelling among sentence sequences. The final article representation $A$ is obtained via sentence attention analogous to eq. \ref{att}.

\noindent \textbf{Final Representation of Case Facts}: Similarly, we pass the combined representation of case facts using pre- and post- interaction similar to Eq. \ref{eq_merge} over a non linear projection, a GRU layer, and sentence level attention to the obtain article-dependent final representation of case facts. To ensure conditioning, we initialize the GRU hidden state with the final representation of the articles $A$. This facilitates capturing the salient case fact information with respect to the specified article.

\subsubsection{Classification Layer}
We pass the article-dependent final representation of the case facts through a nonlinear projection to classify the outcome. 

\subsection{Domain Adaptation Components}
Domain Adaptation aims to make models generalize well from a source to a target domain. Both domains are mapped to a common latent space, reducing differences between their distributions and facilitating domain invariant feature representations. In our case of article-aware COC, we regard reasoning with respect to every ECHR article as a domain and seek to learn article-invariant case facts representations. Put differently, we want our model to learn how to read two texts and interrelate them towards an outcome determination (as lawyers do) with minimal encoding of the information contained in the texts into the model itself. This way, the models can achieve generalization capability to adapt and perform reasoning with regard to articles not seen during training time.

\subsubsection{Domain Discriminator}
We employ a two layer feed forward network as a discriminator which takes the article-dependent case fact representation as input to predict the article (i.e., the domain). We train the discriminator in an adversarial fashion to maximize the model’s ability to capture information required for the outcome task while minimizing its ability to predict the article. This guides the model to generate article-invariant feature representations and improves transferability. Following \cite{ganin2015unsupervised, ganin2016domain}, we perform a min-max game adversary objective optimization using a gradient reversal layer (GRL) between the feature extractor and discriminator. It acts as the identity during the forward pass but, during the backward pass, scales the gradients flowing through by $-\lambda$, making the feature extractor receive the opposite gradients from the discriminator. The overall objective function reduces to:
\begin{equation}
\scalemath{0.8}{
\begin{aligned}
   \arg \min\limits_{\theta_{F}, \theta_{C}, \theta_{D}}     [L_c(C(F(x, a)), y_e) +    \lambda L_d(D(\GRL(F(x, a))), y_a)]
 \end{aligned}}
\end{equation}
where $L_c, L_d$ represents the loss function corresponding to classifier and domain discriminator, respectively, $\lambda$ is the GRL hyperparameter, $x$ is the input, $y_e$ is the outcome  label,  $y_a$ is the class-id of the article, $F$, $C$ and $D$ represents feature extractor, classifier, and discriminator with parameters $\theta_F$, $\theta_C$ and $\theta_D$, respectively. In case of UDA (where the model has access to the text of target domain articles), we discriminate among all the source and target articles. While in case of ADA, we discriminate among source articles only. 

\subsubsection{Wasserstein Method (Distance based)}
Our second method aims to reduce the Wasserstein distance \cite{shen2018wasserstein} between different domain feature distributions. In a given batch, the final feature representations will be fed into the domain critic \cite{arjovsky2017wasserstein}, which is a feedforward network whose output is a single scalar for each batch element. These scalars are then averaged per domain in the batch, resulting in two numbers representing source and target domains, respectively. Their difference can be considered an approximation of the Wasserstein distance between the two feature distributions and becomes the Wasserstein loss component of the network. If the domain critic neural network satisfies the constraint of the Lipschitz-1 continuous function, we calculate the approximate empirical Wasserstein distance by maximizing the following domain critic loss: 

\begin{equation}
\scalemath{0.8}{
\begin{aligned}
    L(X_p, X_q ) =   \frac{1}{n_p} \sum_{x_p \in Xp} f_w(F(x_p)) -  
     \frac{1}{n_q} \sum_{x_q \in X_q} f_w(F(x_q))
\end{aligned}}
\end{equation}
\noindent where $f_w$, $F$ denote the Wasserstein domain critic and feature extractor, respectively, $X_p$ and $X_q$ denote datasets from two domains $p$ and $q$ with $n_p$ and $n_q$ samples, respectively. 

During optimization, a gradient reversal layer \cite{ganin2016domain} between the feature extractor  and domain critic ensures that (a) the domain critic weights are updated such that the Wasserstein loss becomes maximal, while the encoder weights are updated towards minimizing it. Through this procedure, we encourage the model to learn feature representations that are invariant to the covariate shift between the source and the target domain. Since the Wasserstein distance is continuous and differentiable everywhere, we can train the domain critic end-to-end. In case of UDA, we minimize the distance between the source and the target domains, while in case of ADA, we minimize among the different source domains. To enforce the Lipschitz constraints, we clip the weights of the domain critic within a compact space $[-c, c]$ after each gradient update following \cite{arjovsky2017wasserstein}.

\section{Experiments \& Discussion}
\subsection{Baseline}
For the \textbf{fact classification variant}, we employ an architecture similar to the article-aware prediction model but reduced to the case fact based encoding without the interaction mechanism. The output layer is modified to 10 classes and trained against a multi-hot target vector using a binary cross entropy loss. Notably, we freeze the weights in the LegalBERT sentence encoder, both to save computational resources and to reduce the model's susceptibility to shallow surface signals and ensure the comparability of our domain adaptation methods. We describe the detailed hyperparameters for the article-aware prediction model in Appendix Sec. \ref{sec:appendix}

\begin{table}
\centering
\caption{Fact Classification vs Article-aware prediction Performance on Task A and Task B. mic. and mac. indicates micro-F1 and macro-F1 scores, respectively.}
\scalebox{0.8}{
    \begin{tabular}{|p{3cm}||p{1cm}|p{1cm}||p{1cm}|p{1cm}|}
  \hline
   &  \multicolumn{2}{c||}{\textbf{Task B}} & \multicolumn{2}{c|}{\textbf{Task A}}\\
  \hline
   \textbf{Model} & \textbf{mac.} &\textbf{mic.} & \textbf{mac.} &\textbf{mic.} \\ 
 \hline
 Fact Classification & 71.96 & 77.40 & 61.21 & 72.21  \\ 
 Article-aware pred. &  \textbf{74.14} &  \textbf{78.49} & \textbf{67.09} & \textbf{74.77} \\ \hline
  \end{tabular}}
  \label{tab0}
\end{table}

\begin{table*}
\centering
\caption{Task B F1 performance of baseline and domain adaptation models}
    \scalebox{0.75}{
   \begin{tabular}{|p{1.5cm}||p{2.5cm}||p{1cm}|p{1cm}|p{1cm}|p{1cm}||p{1cm}|p{1cm}|p{1cm}|p{1cm}|}
   \hline
   & &  \multicolumn{4}{c||}{\textbf{Transfer $0 \rightarrow 1$}} & \multicolumn{4}{c|}{\textbf{Transfer $0 \leftarrow 1$}}\\
  \hline
    & &  \multicolumn{2}{c|}{\textbf{source : $split\_0$}} & \multicolumn{2}{c||}{\textbf{target : $split\_1$}} & \multicolumn{2}{c|}{\textbf{source : $split\_1$}} & \multicolumn{2}{c|}{\textbf{target : $split\_0$}}\\
  \hline
   \textbf{Setting} & \textbf{Model} & \textbf{mac.} &\textbf{mic.} & \textbf{mac.} &\textbf{mic.} & \textbf{mac.} &\textbf{mic.} & \textbf{mac.} &\textbf{mic.} \\ 
 \hline
  Baseline & Source only & 73.45 & 75.63 & 7.32 & 7.37 & 70.26 & 77.10 & 8.49 & 9.08 \\
  \hline
  \multirow{2}{*}{UDA} & Domain Disc. & \textbf{73.81} &  \textbf{76.95} &  \textbf{13.92} &  14.94  & \textbf{70.63} & \textbf{77.43} & \textbf{22.50} & 26.27\\
   & Wasserstein &  69.63 &  74.86 & 13.17 &  \textbf{18.16} & 66.89 &  75.21 &  20.78 & \textbf{30.30}\\
  \hline
  \multirow{2}{*}{ADA} & Domain Disc. & \textbf{73.76} & \textbf{76.13} & \textbf{9.62} & \textbf{10.77} & \textbf{69.71} & \textbf{76.85} & \textbf{9.30} & \textbf{10.45} \\
   & Wasserstein & 70.17 & 74.80 & 9.14 & 9.89 & 67.46 & 75.25 & 9.26 & 10.38 \\
  \hline
  \end{tabular}}
  \label{tab1B}
\end{table*}

\subsection{Does Article-aware Classification Perform Better than Fact-only Classification?}
Micro-F1 and macro-F1 scores for both tasks A and B with regard to the 10 target articles are given in Table \ref{tab0}. The article-aware model performs better than fact-only classification across the board. In particular, we notice a greater improvement in the macro-F1 score, indicating the article-aware classification approach helps the model to improve performance for sparser articles which are not prominently represented in the case distribution. 
We conjecture that this performance difference can be explained with article-aware classification being subjected to a different training regime. In fact-only classification, a given case's fact text will always be associated with the same multi-hot outcome vector. By contrast, in the fact-aware setting, it will occur multiple times alongside different article texts and the model is forced to predict a single binary outcome variable. This seems to lead the model away from shallow signals towards capturing fact-article correspondence, resulting in a better model. Additionally, the beneficial effect is greater for the harder task of violation classification.

\subsection{Does Domain Adaptation Help to Improve Zero Shot Transferability ?}
We evaluate UDA and ADA on both Task A and Task B with the two article splits. A baseline \textit{source only} model is trained without domain adaptation using the labelled source data only and tested on the target test data directly. Tables \ref{tab1B} and \ref{tab1A} show the performance of different models with our two splits on task B and A, respectively.

\noindent \textbf{{\em Baseline} vs {\em Domain Adaptation}:} From both tables, we observe that the performance of the {\em source only} model on target data is lower compared to their domain adaptation counterparts with a significant margin.  
This indicates that, intuitively, models trained on source data without any adaptation do not generalize to unseen articles. This also highlights the need to have domain adaptation components to achieve a generalizable model. 

\noindent \textbf{UDA:} Under unsupervised domain adaptation, we observe that the Wasserstein distance method performs better on target data than the Domain Discriminator in micro-F1 by a significant margin. It also improved macro-F1 marginally in Task A target data, but is inferior in Task B. Most strikingly, however, Wasserstein performance on source data is lower than the source only baseline across the board, especially with respect to macro-F1. These observations also indicate that the Wasserstein distance method is able to transfer well to certain  articles more than others. This can be attributed to the method influencing feature representations towards a reduction of the mean difference across articles. The distribution of target articles which are closer to the source articles distributions might have gained well. We further validate this hypothesis using an experiment illustrated in sec \ref{sec_related}. On source data, the Domain Discriminator performed better than the source only model, albeit by very small margins but consistent across the tables.  

\noindent \textbf{ADA:} 
On target data, both the Domain Discriminator and Wasserstein distance are comparable across the tables in both metrics. With respect to source data, in task B, the Domain Discriminator performed better than the Wasserstein distance method in both micro and macro F1. Strikingly, in Task A, Wasserstein performance on source data picks up in micro-F1 (slightly even better than source only baseline) but stays behind in macro-F1.

\noindent \textbf{ADA vs UDA:} Unsurprisingly, the performance on target data under ADA tends to be lower compared to UDA due to no access to target article information in this setting compared to UDA. 

The absolute performance levels on the target data immediately suggest that the zero-shot transfer task we propose is very difficult and the discrepancy of performance between source and target data is still large, even in the case of domain adaptation components. This indicates ample opportunity for further research on neural models capable of reasoning with legal text in a way that transfers well to unseen legal domains. Some of the source-target performance divergence can likely be attributed to the model falling prey to spurious correlations that exist in the data, which is especially prominent in the ECtHR datasets that suffer from fact statements not being finalized until the case outcome is known (see our prior work in \citealt{santosh2022deconfounding}. Given this limitation, our zero-shot framework serves as a challenging benchmark in the development of legal NLP models that learn to interrelate case facts and legal source text towards supporting domain experts.



\begin{table*}
\centering
\caption{Task A F1 performance of baseline and domain adaptation models}
    \scalebox{0.75}{
   \begin{tabular}{|p{1.5cm}||p{2.5cm}||p{1cm}|p{1cm}|p{1cm}|p{1cm}||p{1cm}|p{1cm}|p{1cm}|p{1cm}|}
   \hline
   & &  \multicolumn{4}{c||}{\textbf{Transfer $0 \rightarrow 1$}} & \multicolumn{4}{c|}{\textbf{Transfer $0 \leftarrow 1$}}\\
  \hline
    & &  \multicolumn{2}{c|}{\textbf{source : $split\_0$}} & \multicolumn{2}{c||}{\textbf{target : $split\_1$}} & \multicolumn{2}{c|}{\textbf{source : $split\_1$}} & \multicolumn{2}{c|}{\textbf{target : $split\_0$}}\\
  \hline
   \textbf{Setting} & \textbf{Model} & \textbf{mac.} &\textbf{mic.} & \textbf{mac.} &\textbf{mic.} & \textbf{mac.} &\textbf{mic.} & \textbf{mac.} &\textbf{mic.} \\ 

 \hline
  Baseline & Source only & 63.62 &  71.98 &  3.14 & 3.78 & 67.79 & 74.57 & 5.80 & 8.02 \\
  \hline
  \multirow{2}{*}{UDA} & Domain Disc. & \textbf{64.65} & \textbf{72.52} & 9.52 & 9.87 & \textbf{68.19} & \textbf{75.32} & 14.47 & 16.51\\
   & Wasserstein & 60.26 & 71.46 & \textbf{11.04} & \textbf{18.20} & 63.56 & 74.89 & \textbf{15.23} & \textbf{26.20}\\
  \hline
  \multirow{2}{*}{ADA} & Domain Disc. &  \textbf{64.89} & 72.08 &  7.18 &  \textbf{7.78}  & \textbf{67.12} & 74.43 & 6.45 & 9.34\\
   & Wasserstein & 61.78 & \textbf{72.36} & \textbf{7.27} & 7.61 & 65.71 & \textbf{74.88} & \textbf{6.71} & \textbf{9.71}\\
   \hline
  \end{tabular}}
  \label{tab1A}
\end{table*}

\subsection{How does Encoder Pre-Training influence Zero-Shot Transferability?}
We conduct an additional experiment on Task A with $split\_1$ as source and $split\_0$ as target, where we replace LegalBERT embeddings used in the encoding layer with BERT base embeddings \cite{kenton2019bert}, and report its performance in Table \ref{tab2}. Comparing it to Transfer $0 \leftarrow 1$ in Table \ref{tab1A}, we observe that the BERT base model performs worse on target data than the LegalBERT encoder. In particular, the best performing Wasserstein domain adaptation model drops from 26.2 to 16.36, much more than the Domain Discriminator. We leave an exploration of this asymmetric effect of the pre-training regime across different domain adaptation strategies to future work.

Base BERT performs similarly on the source domain. This indicates that even a non-legally pre-trained encoder can be harnessed to reach comparable in-domain performance. However, to generalize to unseen target articles, domain specific pre-training is beneficial. It should be noted that LegalBERT \cite{chalkidis2020legal} has been pre-trained on a collection of ECtHR decisions that may include cases from LexGLUE's test partition, thereby possibly injecting domain-specific information about the target articles into the encoding. 

\begin{table}
\centering
\caption{Task A F1 Performance in one split using BERT base embeddings (as opposed to Legal Bert)}
\scalebox{0.73}{
     \begin{tabular}{|p{1cm}||p{1.75cm}||p{0.9cm}|p{0.9cm}||p{0.9cm}|p{0.9cm}|}
  \hline
    & &  \multicolumn{2}{c||}{\textbf{source : $split\_1$}} & \multicolumn{2}{c|}{\textbf{target : $split\_0$}}\\
  \hline
   \textbf{Setting} & \textbf{Model} & \textbf{mac.} &\textbf{mic.} & \textbf{mac.} &\textbf{mic.} \\ 
 \hline
  \multirow{2}{*}{UDA} & Dom. Disc. &  68.01 & 75.26 &  13.68 & 15.21 \\
   & Wasserstein & 62.15 & 74.32 & 14.12 & 16.36 \\
  \hline
  \multirow{2}{*}{ADA} & Dom. Disc. & 67.92 & 75.32 & 4.77 & 7.44 \\
   & Wasserstein & 66.71 &  74.95 & 4.73 & 7.65 \\
   \hline
  \end{tabular}}
  \label{tab2}
\end{table}

\subsection{How does Article Relatedness Affect Zero-Shot Transferability?} \label{sec_related}

To test whether article relatedness between source and target domains affect performance, we experiment with Article P1-1 (Article 1 of  Additional Protocol 1 - The Protection of Property) as the target domain. This simulates the realistic scenario of our zero shot setting where the convention is amended with an additional protocol. We then constructed one related and one unrelated source domain based on the suggestion provided by a legal expert (the second author) while ensuring training sets of similar size. The related domain consists of articles 6 (right to a fair trial) and 8 (right to respect for private and family life). The unrelated domain articles comprise articles 2 (right to life), 3 (prohibition of torture, and 5 (right to liberty and security).

We report the performance on Task A for target P1-1 in Table \ref{tab3}. We observe that the related source domain is able to perform better across the board, confirming the intuition that relatedness between source and target is an important factor to be considered when training a model for transferability. As before, we observe that UDA achieves higher performance overall as it has the chance to see article P1-1 during training. Interestingly, we observe the Wasserstein method outperforming the Domain Discriminator for the related source, but vice versa for the unrelated source. We believe this is owed to related articles forming similar feature distributions and thereby making it easy for the Wasserstein distance to facilitate adaptation. This case study suggests the design of domain adaptation components which derive information more from related articles than unrelated ones when transferring to a target article. This raises a related question of how article relatedness could be determined by the model itself rather than a priori by an expert.

\begin{table}
\centering
\caption{Task A F1 target performance on article P1-1 with related and unrelated source domains}
\scalebox{0.73}{
     \begin{tabular}{|p{1cm}||p{1.75cm}||p{0.9cm}|p{0.9cm}||p{0.9cm}|p{0.9cm}|}
  \hline
    & Source &  \multicolumn{2}{c||}{\textbf{Related}} & \multicolumn{2}{c|}{\textbf{Unrelated}}\\
  \hline
   \textbf{Setting} & \textbf{Model} & \textbf{mac.} &\textbf{mic.} & \textbf{mac.} &\textbf{mic.}  \\ 
 \hline
  \multirow{2}{*}{UDA} & Dom. Disc. & 54.13 &  73.71 &  43.52 & 65.72  \\
   & Wasserstein  &  62.35 &  74.64 & 34.01 & 49.62  \\
  \hline
  \multirow{2}{*}{ADA} & Dom. Disc.  & 42.79 &  68.46 & 37.12 & 56.16   \\
   & Wasserstein  &  43.25 &  69.91 & 26.87 & 38.28 \\
   \hline
  \end{tabular}}
  \label{tab3}
\end{table}

\section{Conclusion}
We cast case outcome classification on ECtHR data into an article-aware architecture. This configuration is inspired by realistic legal reasoning involving both the case facts and convention articles to determine possible allegations/violations. Assuming non-finetuned pre-trained encoders, we observe a performance improvement over a simple fact-only classification model. It also enables us to conduct experiments in zero shot transfer COC with and without access to unlabeled target data during domain adaptation. While we show that domain adaptation techniques are in principle suitable to facilitate generalization, the divergence between source and target domain performance is large and this task variant is very difficult. We further observe that the effectiveness of domain adaptation interacts with law-specific pre-training of transformer-based encoders and with the relatedness of the source and target domains. Overall, this zero-shot COC task formulation opens up new research opportunities towards legal NLP models that are more aligned with expert reasoning.

\section*{Limitations}
We cast the legal judgment prediction task into an article-aware classification setting and create a zero-shot benchmark on a corpus of ECtHR cases. Matching between the text of legal sources and case fact descriptions varies greatly between different legal systems and subdomains, and is highly dependent on the textual nature of the case fact and legal sources. Specific to our context, for example, we have discussed the ECtHR fact statements as being influenced by the eventual case outcome and not suitable for prospective prediction in sec \ref{sec_intro}. COC as article-aware classification in other jurisdictions will likely lead to different levels of task difficulty, absolute performance, and zero shot transferability. In particular, many legal areas require multiple sources to be applied in conjunction to a set of case facts.

Technically, a major hurdle dealing with corpora related to the legal domain is their lengthy nature. We resort to hierarchical models, which are inherently limited in that tokens across long distances cannot directly attend to one another. This restriction of hierarchical models is still underexplored (but see preliminary work in, e.g. \citealt{dai2022revisiting,chalkidis2022exploration}). Additionally, we freeze the weights in the LegalBERT sentence encoder, both to save computational resources and to reduce the model's susceptibility to shallow surface signals and ensure the comparability of our domain adaptation methods, in particular with respect to the impact of domain-specific pre-training. We leave an exploration of COC as article-aware classification with fine-tuned encoders for future work.

\section*{Ethics Statement}
We experiment with a publicly available datasets of ECtHR decisions, which has been derived from the public court database HUDOC\footnote{\url{https://hudoc.echr.coe.int}}. These decisions contain real names of the parties involved without any anonymization. We hence do not consider our experiments to produce any additional harmful effects relating to personal information.

The task of legal judgment prediction raises ethical, civil rights, and legal policy concerns, both general and specific to the European Court of Human Rights (e.g., \cite{fikfak2021future} on system bias and court caseload). The main premise of this work is to make incremental technical progress towards enabling systems to work with case outcome information in a way that is aligned with how human experts analyze case facts through an interplay with complex legal sources. We do not advocate for the practical application of COC/LJP systems by courts, but rather explore how their core functionality of processing legal text can be made as expert-aligned as possible. Our research group is strongly committed to research on such models as a means to derive insight from legal data for purposes of increasing transparency, accountability, and explainability of data-driven systems in the legal domain.

We are conscious that, by adapting pre-trained encoders, our models inherit any biases they contain. Similarly, the ECtHR case collection as historical data may contain a data distribution in which sensitive attributes of individuals (e.g., applicant gender) may have some predictive signal for the allegation/violation variable (see, e.g., \cite{chalkidis2022fairlex}). We believe the results we observe in our COC experiments to not be substantially related to such encoded bias. However, legal NLP systems leveraging case outcome information and intended for practical deployment should naturally be scrutinized against applicable equal treatment imperatives regarding their performance, behavior, and intended use.

All models of this project were developed and trained on Google Colab. We did not track computation hours.

\bibliography{main}
\bibliographystyle{acl_natbib}

\appendix

\section{Implementation Details}
\label{sec:appendix}
We employ a maximum sentence length of 256 and document length (number of sentences) of 50. Our word level attention context vector size is 300. The sentence level GRU encoder dimension is 200 (i.e. 400 bidirectional), and the sentence level attention vector dimension is 200. The entailment classifier hidden layer also has size 200. Domain discriminator and critic have two layered networks with hidden layers of size 200 and 100. The entailment classifier is trained with a binary cross entropy loss while the domain discriminator is trained with cross entropy loss over a one-hot domain vector. The model is optimized end-to-end using Adam \cite{kingma2014adam}. The dropout rate \cite{srivastava2014dropout} in all layers is 0.1. To handle data skewness in the entailment setup, we employ a custom batch sampler which ensures every batch contains 4 different articles as well as 2 positive and 2 negative instances per article. Our batch size is 16. We employ a learning rate scheduler based on loss plateau decay. For adversarial training using GRL, following \cite{ganin2015unsupervised}, we set the $\lambda$ in gradient reversal to be  $\lambda = \frac{2}{1+\exp(-\gamma p)} -1 $ where $p = \frac{t}{T}$, where t and T denote current training step and total training steps. $\gamma$ is determined using a grid search over [0.05, 0.1, 0.15, 0.2]. We employ a 10 class domain discriminator (5 from source and 5 from target) in the case of UDA and a 5 class discriminator in the case of ADA. We reduce the mean between instances of a particular article of source and target in the case of UDA. In the case of ADA, we reduce the mean between instances of different articles in the source domain.

\end{document}